\documentclass[conference]{IEEEtran}
\IEEEoverridecommandlockouts
\usepackage{cite}
\usepackage{amsmath,amssymb,amsfonts}
\usepackage{algorithmic}
\usepackage{graphicx}
\usepackage{textcomp}
\usepackage{xcolor}
\usepackage{hyperref}
\usepackage{url}
\usepackage{multirow}
\usepackage{comment}
\usepackage{balance}

\def\BibTeX{{\rm B\kern-.05em{\sc i\kern-.025em b}\kern-.08em
    T\kern-.1667em\lower.7ex\hbox{E}\kern-.125emX}}
\begin{document}

\title{An Experimental Study on Generating Plausible Textual Explanations for Video Summarization*\\
\thanks{This work was supported by the EU Horizon Europe programme under grant agreement 101070109 TransMIXR. \\ IEEE CBMI 2025. \copyright IEEE. This is the authors' accepted version. The final publication is available at https://ieeexplore.ieee.org/}
}

\author{\IEEEauthorblockN{Thomas Eleftheriadis}
\IEEEauthorblockA{
\textit{ITI, CERTH}\\
Thessaloniki, Greece \\
thomelef@iti.gr}
\and
\IEEEauthorblockN{Evlampios Apostolidis}
\IEEEauthorblockA{
\textit{ITI, CERTH}\\
Thessaloniki, Greece \\
apostolid@iti.gr}
\and
\IEEEauthorblockN{Vasileios Mezaris}
\IEEEauthorblockA{
\textit{ITI, CERTH}\\
Thessaloniki, Greece \\
bmezaris@iti.gr}
}

\maketitle

\begin{abstract}
In this paper, we present our experimental study on generating plausible textual explanations for the outcomes of video summarization. For the needs of this study, we extend an existing framework for multigranular explanation of video summarization by integrating a SOTA Large Multimodal Model (LLaVA-OneVision) and prompting it to produce natural language descriptions of the obtained visual explanations. Following, we focus on one of the most desired characteristics for explainable AI, the plausibility of the obtained explanations that relates with their alignment with the humans' reasoning and expectations. Using the extended framework, we propose an approach for evaluating the plausibility of visual explanations by quantifying the semantic overlap between their textual descriptions and the textual descriptions of the corresponding video summaries, with the help of two methods for creating sentence embeddings (SBERT, SimCSE). Based on the extended framework and the proposed plausibility evaluation approach, we conduct an experimental study using a SOTA method (CA-SUM) and two datasets (SumMe, TVSum) for video summarization, to examine whether the more faithful explanations are also the more plausible ones, and identify the most appropriate approach for generating plausible textual explanations for video summarization.
\end{abstract}


\section{Introduction}
Video summarization technologies aim to generate a short summary of the full-length video by selecting the most informative and important frames (key-frames) or fragments (key-fragments) of it. The current state of the art is represented by methods that rely on the use of trained deep learning network architectures \cite{9594911}. Nevertheless, after their training, these methods are used as ``black-box'' systems, since the provision of explanations about the video summarization results is not an inherent feature of the employed deep network architectures. The generation of such explanations, that could significantly facilitate and advance media content production \cite{apostolidis_chapter}, has been investigated only to a small extent thus far. 

A first attempt to formulate the task of explainable video summarization was made in \cite{10019643} and extended in \cite{10.1145/3607540.3617138}, where the authors assessed the performance of various attention-based explanation signals that had been previously used to form explanations for network architectures dealing with tasks from the NLP domain \cite{chrysostomou-aletras-2021-improving, pmlr-v162-liu22i}. Another approach that provides clues about the decisions of the video summarization network with the help of causality graphs between input data, output scores, summarization criteria and data perturbations, was presented in \cite{10208771}. Finally, an integrated framework for multi-granular explanation of video summarization, that indicates the video fragments and visual objects which influenced the most the output of the summarization network, was described in \cite{10.3389/frsip.2024.1433388}. Despite the fact that the methods above provide useful insights about the decision mechanism of video summarization networks, the produced graph-based or fragment- and object-level explanations require interpretation by a human expert. 

In this paper, we extend the framework for multi-granular explanation of video summarization from \cite{10.3389/frsip.2024.1433388}, in order to also provide textual explanations. Using the extended framework (depicted in Fig. \ref{fig:overview}), we perform an experimental study on generating plausible textual explanations for video summarization, that require no human interpretation. Plausibility is among the suggested human-grounded criteria to evaluate explanations of networks decisions \cite{Salih2024}. According to the relevant literature \cite{jacovi-goldberg-2020-towards, jin2024plausibility}, plausibility measures how much aligned the explanation is with the human expectation. In this sense, it provides an indication of how reasonable and convincing the explanation is, to humans. Nevertheless, a plausible explanation is not necessarily a faithful one, and vice-versa. For example, using the obtained visual explanations (in the form of heatmaps) after a Fisher Vector-based classifier correctly classifies images of the class ``horse'', Lapuschkin et al. \cite{Lapuschkin2019} observed that the classifier based its decisions on the presence of a copyright watermark on one of the corners of the images, that accidentally persisted in this very popular Pascal VOC dataset \cite{Everingham2010}. The acquired visual explanations were proven to be highly faithful, since the removal of this watermark from images showing horses resulted in a different classification outcome. However, they had nothing to do with the humans' expectations, and thus they were not plausible at all. 

In the context of explainable video summarization, we argue that a plausible explanation should focus on objects and events of the video that either appear in the video summary or exhibit a strong semantic relationship with its content. Based on this argumentation and the extended framework for explainable video summarization, we quantify the plausibility of the obtained visual-based explanation for the video summarization output (i.e., the video summary), by computing the semantic overlap between the associated textual explanation and the textual description of the video summary. For this, we represent both the textual explanation and the textual description of the video summary using two SOTA methods for creating meaningful sentence embeddings, namely SBERT \cite{reimers-gurevych-2019-sentence} and SimCSE \cite{gao-etal-2021-simcse}, and compute the cosine similarity of each pair of representations. Building on the above, we aim to answer the following research questions:
\begin{itemize}
    \item When the explanation of the video summarization is a condensed one, formed only by the most influential video fragment according to an explanation method, is the most faithful explanation (based on objective evaluation measures, such as Discoverability+ (Disc+) \cite{10019643, 10.3389/frsip.2024.1433388}) also the most plausible one?
    \item When the explanation of the video summarization is a more detailed one, formed using the three most influential video fragments according to an explanation method, is the most faithful explanation also the most plausible one? And what is the optimal approach for producing the textual explanation in this case?
\end{itemize}

Our contributions are as follows:
\begin{itemize}
    \item We extend an integrated framework for explainable video summarization \cite{10.3389/frsip.2024.1433388}, by introducing the SOTA Large Multimodal Model LLaVA-OneVision \cite{li2024llavaonevisioneasyvisualtask} and prompting it to produce textual descriptions of the obtained fragment-level visual explanations.
    \item We specify the characteristics of plausible explanations in the context of explainable video summarization, and propose to evaluate the plausibility of visual explanations by computing the semantic overlap between their textual descriptions and the textual descriptions of the corresponding video summaries, using two SOTA methods for creating sentence embeddings.
    \item Based on the extended framework and the proposed plausibility evaluation approach, we conduct an experimental study aiming to examine whether the more faithful explanations are also the more plausible ones, and identify the most appropriate approach for generating plausible textual explanations for video summarization.
\end{itemize}

\section{Related Work}

Over the last years there is a growing interest of researchers in developing methods that provide explanations about the output of neural networks dealing with various video analysis tasks, such as video classification \cite{10.1007/978-3-030-69541-5_25, Li2021TowardsVE, 10539635}, action classification and reasoning \cite{10.1145/3343031.3351040, HAN2022212}, video activity recognition \cite{Roy2019ExplainableAR}, video anomaly detection \cite{278c9656de614a479c93c6dead189ff4, 10205367, 9468958, 9706981}, and deepfake video detection \cite{10350382, 10677972, 10972511}. With respect to explainable video summarization, a first attempt was made in \cite{10019643, 10.1145/3607540.3617138}, where the task was formulated as the production of an explanation mask indicating the parts of the video that influenced the most the estimates of a video summarization network about the frames' importance. The performance of various attention-based explanation signals was evaluated using a SOTA network architecture (called CA-SUM \cite{10.1145/3512527.3531404}) and two datasets for video summarization (SumMe \cite{10.1007/978-3-319-10584-0_33} and TVSum \cite{7299154}), by investigating the network’s input-output relationship for different input replacement functions, and using a set of tailored evaluation measures. Following a different approach, Huang et al. \cite{10208771} described a method that leverages ideas from Bayesian probability and causation modeling, and provides a form of explanation using causality graphs that show relations between input data, output importance scores, summarization criteria and applied perturbations. However, the produced graphs require interpretation by a human expert, while the performance of these explanations was not evaluated through quantitative or qualitative analysis. Finally, building on \cite{10019643, 10.1145/3607540.3617138}, Tsigos et al. \cite{10.3389/frsip.2024.1433388} presented an integrated framework for multi-granular explanation of video summarization, that can be applied both on ``white-box'' (transparent to the user) and ``black-box'' models, and produces visual explanations indicating the video fragments and visual objects that influenced the most, the output of the video summarization network.

Our work is most closely related to methods dealing with the explanation of the video summarization results \cite{10019643, 10.1145/3607540.3617138, 10208771, 10.3389/frsip.2024.1433388}. However, contrary to these works, that produce \textbf{visual explanations} in the form of causality graphs illustrating relations between input data, output scores, summarization criteria and applied perturbations \cite{10208771}, or in the form of short videos containing the most influential video fragments and keyframe collections highlighting the most influential visual objects \cite{10.3389/frsip.2024.1433388}, we aim to provide the users with \textbf{plausible textual explanations} for the video summarization results. The provision of such explanations eliminates the need for human interpretation, thus allowing more immediate understanding of the decision mechanism of video summarization networks.

\section{Proposed Explanation Approach}

\begin{figure*}[t]
\centerline{\includegraphics[width=\textwidth]{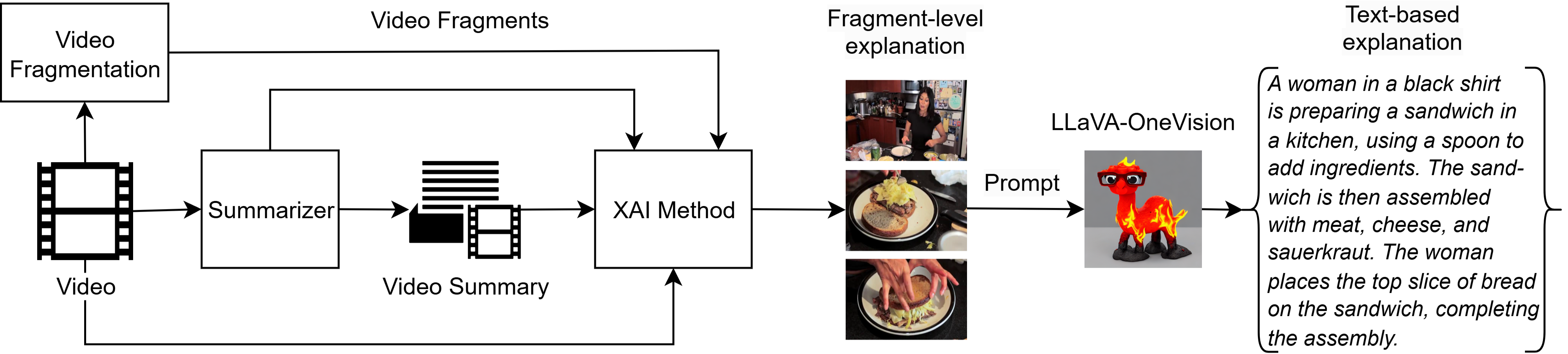}}
\caption{A high-level overview of the extended framework for producing textual explanations for the video summarization results.}
\label{fig:overview}
\end{figure*}

A high-level overview of the extended framework for producing textual explanations for the video summarization results, is given in Fig \ref{fig:overview}. The basis for our developments was the integrated framework for multi-granular explanation of video summarization, from \cite{10.3389/frsip.2024.1433388}. In the core of this framework there is an explanation method that gets as input the full-length video, the summarization model and the produced video summary (formed by the three top-scoring video fragments by the summarizer). Depending on the access to the internal structure of the summarization model, the explanation method can be either model-agnostic (such as the post-hoc perturbation-based LIME method \cite{ribeiro2016should}), or model-specific (such as the attention-based method from \cite{10.1145/3512527.3531404}). After the end of the analysis, the framework produces three different types of visual explanations; i.e., a fragment-level explanation indicating the video fragments that influenced the most the generation of the video summary, and two object-level explanations showing the most influential visual objects within the fragment-level explanation and the video summary, respectively.

The aforementioned visual explanations provide cues about the pieces of the visual content that affected the decisions of the video summarization network. Nevertheless, the extraction of a coherent and human-understandable explanation requires the observation and interpretation of the provided visual cues. To facilitate the acquisition of such an explanation, we worked towards producing textual explanations for the video summarization results. For this, we extended the aforementioned framework, by integrating a pretrained model of the LLaVA-OneVision Large Multimodal Model \cite{li2024llavaonevisioneasyvisualtask} and prompting it to generate natural language descriptions of the produced fragment-level explanations. LLaVA-OneVision is an open-source, multimodal vision-language model that has been designed to process and generate text from images, videos and multiple images. It has been trained following a multi-step training approach and using: a) a single-image dataset collection containing various categories of data such as math/reasoning and OCR (called Single-image 3.2M in \cite{li2024llavaonevisioneasyvisualtask}), and b) a single-image, multi-image and video dataset collection (called OneVision 1.6M in \cite{li2024llavaonevisioneasyvisualtask}). Given the fact that LLaVA-OneVision was not trained using images that resemble the masked keyframes of the produced object-level explanations by the utilized explanation framework \cite{10.3389/frsip.2024.1433388}, we decided to use only the fragment-level explanations as input to the integrated LLaVA-OneVision model in the extended framework. Using tailored prompts, the integrated LLaVA-OneVision model generates natural language descriptions of the visual content of the aforementioned explanations, forming textual explanations for the video summarization results.

\section{Experiments}

\subsection{Evaluation approach}
\label{subsec:eval_approach}

\begin{figure*}[t]
\centerline{\includegraphics[width=\textwidth]{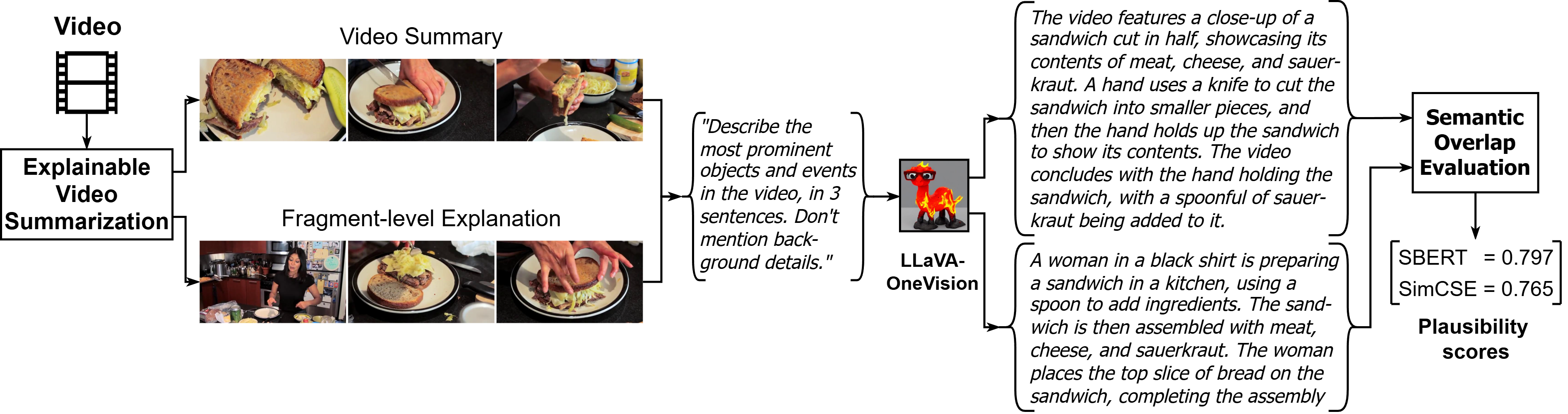}}
\caption{An overview of the proposed approach for evaluating the plausibility of visual explanations.}
\label{fig:plausibility_eval}
\end{figure*}

To evaluate the faithfulness of visual explanations, we focused on the top-k scoring fragments (with k equal to $1$ and $3$) by an explanation method and employed the Disc+ measure (one of the measures of the evaluation protocol introduced in \cite{10019643} and used also in \cite{10.3389/frsip.2024.1433388}). Disc+ assesses if the top-k scoring fragments by the explanation method have a significant impact to the network's output. For a given video, it is computed as $\Delta E(\boldsymbol{X},\boldsymbol{\hat{X}}) = \tau(\boldsymbol{y}, \boldsymbol{\hat{y}})$, where $\boldsymbol{X}$ and $\boldsymbol{\hat{X}}$ are the original and the updated frame representations after masking out the frames of the top-k scoring fragments, respectively, $\boldsymbol{y}$ and $\boldsymbol{\hat{y}}$ are the outputs of the summarization network for $\boldsymbol{X}$ and $\boldsymbol{\hat{X}}$, and $\tau$ is the Kendall's correlation coefficient \cite{kendall1945treatment}. $\Delta E$ ranges in $[-1, +1]$; values close to $+1$ indicate strong agreement between $\boldsymbol{y}$ and $\boldsymbol{\hat{y}}$ (thus, a minor impact after a perturbation), and values close to $-1$ signify strong disagreement between $\boldsymbol{y}$ and $\boldsymbol{\hat{y}}$ (thus, a major impact after a perturbation). So, the lower the Disc+ measure is, the greater the ability of the explanation method to identify the video fragments with the highest influence to the video summarization network.

To evaluate the plausibility of visual explanations, we rely on the concept that a plausible, and thus reasonable and convincing, explanation should contain visual objects and events of the video that are either present in the video summary or semantically relevant with its content. Otherwise, i.e., if the explanation includes objects or events that are completely irrelevant to the parts of the video that formulate the summary, it will be weakly aligned with the humans' expectation and thus won't be plausible. One way to estimate the semantic relevance between the video summary and the visual explanation is the use of similarity-based assessments; however, visual similarity does not necessarily imply semantic relevance, and vice-versa. Instead of this, we quantify this relevance, and thus the plausibility of a visual explanation, by utilizing the generated textual description and computing its semantic overlap with the textual description of the video summary, as depicted in Fig. \ref{fig:plausibility_eval}. More specifically, we generate a textual description of the video summary by concatenating its fragments in temporal order and then prompting the integrated LLaVA-OneVision model to describe the formulated video, as we did to obtain the textual description of the visual explanation. Then, we represent both of the acquired textual descriptions using two SOTA methods for creating sentence embeddings, namely SBERT \cite{reimers-gurevych-2019-sentence}\footnote{\url{https://huggingface.co/sentence-transformers/all-mpnet-base-v2}} SimCSE \cite{gao-etal-2021-simcse}\footnote{\url{https://huggingface.co/princeton-nlp/sup-simcse-bert-base-uncased}}, and compute the cosine similarity of each pair of representations. The computed scores range in $[0,+1]$, with higher scores indicating greater plausibility.

\subsection{Utilized datasets}

Following \cite{10.3389/frsip.2024.1433388}, we conducted our experimental study using the SumMe \cite{10.1007/978-3-319-10584-0_33} and TVSum \cite{7299154} benchmarking datasets for video summarization. SumMe contains $25$ videos with varying visual content that have been captured from both first-person and third-person view and are up to $6$ minutes long. TVSum includes $50$ videos from $10$ categories of the TRECVid MED dataset that are up to $11$ minutes long.

\subsection{Implementation details}

Videos were downsampled to $2$ fps and deep representations of the sampled frames were obtained from the pool5 layer of a pre-trained model of GoogleNet \cite{7298594} on the ImageNet \cite{5206848} dataset. Video summarization was performed using pre-trained models of the CA-SUM method \cite{10.1145/3512527.3531404} on the SumMe and TVSum datasets\footnote{\url{https://github.com/IDT-ITI/XAI-Video-Summaries}}. Following \cite{10.3389/frsip.2024.1433388}, we set the number of applied perturbations $M$ for producing explanations equal to $20,000$. For generating the textual explanations, we employed the 7B-parameters version of the LLaVA-OneVision model\footnote{\url{https://huggingface.co/llava-hf/llava-onevision-qwen2-7b-ov-hf}} optimized with 4-bit quantization, keeping ``temperature'' equal to zero to get more deterministic responses. Moreover, we examined various prompts ($10$ for each of the approaches presented in Section \ref{subsec:quan_results}) to conclude to the used ones. All experiments were carried out on an NVIDIA RTX 4090 GPU. The code and data for reproducing our experiments are publicly-available at: \url{https://github.com/IDT-ITI/Text-XAI-Video-Summaries}.

\subsection{Quantitative results}
\label{subsec:quan_results}

Initially, we aimed to investigate whether the most faithful explanation (according to the Disc+ measure) is also the most plausible one, when the visual explanation is formed by the top-scoring video fragment by an explanation method. For this, we took into account the attention-based and the LIME-based explanation methods that were used in \cite{10.3389/frsip.2024.1433388}, and evaluated the faithfulness of the obtained visual explanations using the set of videos from \cite{10.3389/frsip.2024.1433388} with at least one top-scoring fragment by these methods (Video Set 1). The results of this assessment, reported in the upper part of Table \ref{tab:faithfulness}, are in line with the findings in \cite{10.3389/frsip.2024.1433388} and indicate the capacity of the attention-based explanation method to produce more faithful explanations compared to the LIME-based method, on both datasets. Then, we assessed the plausibility of the obtained visual explanations on the same set of videos (Video Set 1) after prompting LLaVA-OneVision to describe the content of each explanation as follows: \textit{``Describe the most prominent objects and events in the video, in 3 sentences. Don't mention background details.''} (Approach 1). The results presented in the upper part of Table \ref{tab:plausibility}, show that the LIME-based explanation method leads to more plausible explanations in both datasets and according to both of the utilized sentence embedding methods. Such a finding indicates that \textbf{in the case of condensed visual explanations - formed by the single most influential video fragment for the video summarization network - higher faithfulness does not necessarily imply higher plausibility.}

\begin{table}[t]
\caption{Faithfulness (in terms of Disc+) of the obtained visual explanations for two different sets of videos of the SumMe and TVSum datasets, that have at least one (Video Set 1) and three (Video Set 2) top-scoring fragments by the explanation methods. The best (lower) scores are shown in bold.}
\begin{center}
\begin{tabular}{|l|cc|}
\hline
\multirow{2}{*}{} & \multicolumn{1}{c|}{\begin{tabular}[c]{@{}c@{}}Attention-based\\ explanation\end{tabular}} & \begin{tabular}[c]{@{}c@{}}LIME-based\\ explanation\end{tabular} \\ \cline{2-3} 
                  & \multicolumn{2}{c|}{Video Set 1}                                                                                                                              \\ \hline
SumMe             & \multicolumn{1}{c|}{\textbf{0.579}}                                                        & 0.787                                                            \\
TVSum             & \multicolumn{1}{c|}{\textbf{0.565}}                                                        & 0.807                                                            \\ \hline
                  & \multicolumn{2}{c|}{Video Set 2}                                                                                                                              \\ \hline
TVSum             & \multicolumn{1}{c|}{\textbf{0.483}}                                                        & 0.615                                                            \\ \hline
\end{tabular}
\end{center}
\label{tab:faithfulness}
\end{table}

\begin{figure*}[t]
\centerline{\includegraphics[width=\textwidth]{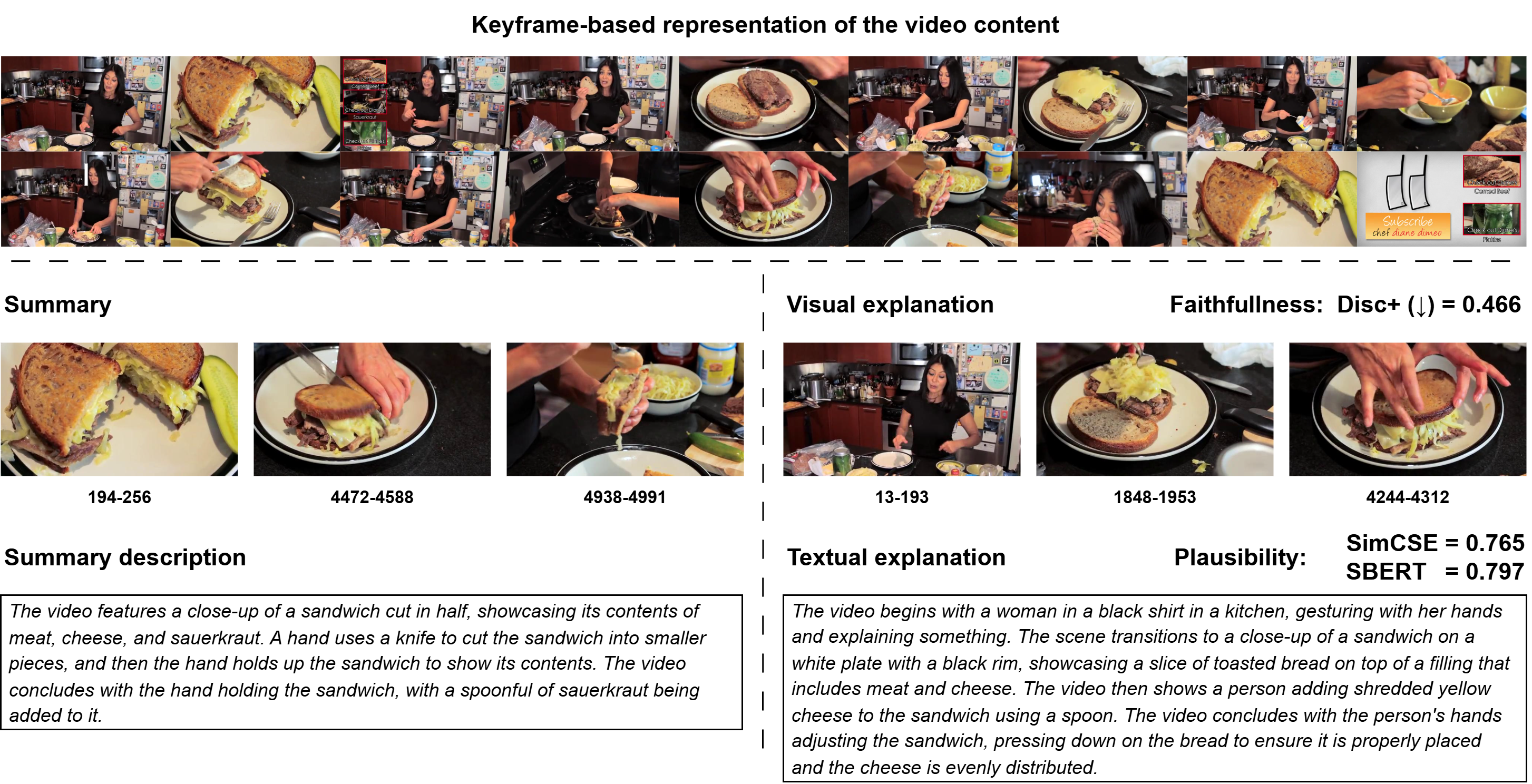}}
\caption{Top: keyframe-based representation of the content of a TVSum video, titled ``Reuben Sandwich with Corned Beef \& Sauerkraut''. Middle: keyframe-based representations of the video summary and the produced fragment-level explanation by the attention-based explanation method. Bottom: textual descriptions of the video summary and the fragment-level explanation obtained by the integrated LLaVA-OneVision model, and the computed plausibility scores.}
\label{fig:qual_ex1}
\end{figure*}

\begin{table}[t]
\caption{Plausibility of the obtained visual explanations for two different sets of videos of the SumMe and TVSum datasets, that have at least one (Video Set 1) and three (Video Set 2) top-scoring fragments by the explanation methods. The best (higher) scores of each row are shown in bold.}
\begin{center}
\begin{tabular}{|l|cccc|}
\hline
\multirow{2}{*}{} & \multicolumn{2}{c|}{\begin{tabular}[c]{@{}c@{}}Attention-based\\ explanation\end{tabular}} & \multicolumn{2}{c|}{\begin{tabular}[c]{@{}c@{}}LIME-based\\ explanation\end{tabular}} \\ \cline{2-5} 
                  & \multicolumn{4}{c|}{Video Set 1}                                                                                                                                                   \\ \hline
                  & \multicolumn{1}{c|}{SBERT}                   & \multicolumn{1}{c|}{SimCSE}                 & \multicolumn{1}{c|}{SBERT}                           & SimCSE                         \\ \hline
SumMe - Approach 1            & \multicolumn{1}{c|}{0.713}                   & \multicolumn{1}{c|}{0.676}                  & \multicolumn{1}{c|}{\textbf{0.729}}                  & \textbf{0.712}                 \\
TVSum - Approach 1            & \multicolumn{1}{c|}{0.535}                   & \multicolumn{1}{c|}{0.532}                  & \multicolumn{1}{c|}{\textbf{0.593}}                  & \textbf{0.577}                 \\ \hline
                  & \multicolumn{4}{c|}{Video Set 2}                                                                                                                                      \\ \hline
TVSum - Approach 1        & \multicolumn{1}{c|}{\textbf{0.617}}                   & \multicolumn{1}{c|}{\textbf{0.624}}                  & \multicolumn{1}{c|}{0.588}                           & 0.589                          \\
TVSum - Approach 2        & \multicolumn{1}{c|}{\textbf{0.626}}          & \multicolumn{1}{c|}{\textbf{0.630}}         & \multicolumn{1}{c|}{0.622}                  & 0.629                 \\ \hline
\end{tabular}
\end{center}
\label{tab:plausibility}
\end{table}

Next, we examined whether our previous finding holds true also in the case of more detailed visual explanations. For this, we employed the subset of videos from \cite{10.3389/frsip.2024.1433388} with at least three top-scoring video fragments by the explanation methods (Video Set 2). In this case the three fragments are concatenated in temporal order, forming a video that is then described using the LLaVA-OneVideo and the same prompt as before. The results about the explanations' faithfulness and plausibility are reported in the lower part of Table \ref{tab:faithfulness} and the third row of Table \ref{tab:plausibility}, respectively. These results demonstrate that the most faithful visual explanations (obtained once again by the attention-based method) are also the most plausible ones. So, it seems that increasing the number of video fragments that form the visual explanation leads to a different conclusion about the alignment between faithful and plausible explanations. Contrary to our previous finding, \textbf{in the case of more descriptive visual explanations - formed by the three most influential video fragments for the video summarization network - the most faithful explanation is also the most plausible one}. Moreover, a comparison across the different sets of videos from the TVSum dataset (i.e., Video Set 1 and 2), shows that this increase leads to visual explanations that are more faithful (as indicated by the lower scores in the third row of Table \ref{tab:faithfulness}) and more plausible (as denoted by the mostly higher scores in the third row of Table \ref{tab:plausibility}), compared to the more condensed ones.

Finally, we tried to find out what is the optimal approach for producing the textual explanation in the case of more detailed and descriptive visual explanations. So, in addition to our initial methodology described above (see Approach 1), we considered another more fine-grained one. In this case, the initial methodology is applied on each different fragment of the visual explanation and the obtained descriptions (three in total) are then summarized by LLaVA-OneVision, using the following prompt: \textit{``Write a brief summary that covers all 3 descriptions equally. Avoid assumptions and background details.''} (Approach 2). The results of this evaluation, reported in the last row of Table \ref{tab:plausibility}, show that second approach leads to constantly higher plausibility scores. The increase is more pronounced for the LIME-based method, but the attention-based method is again the best-performing one according to both sentence embedding methods. These findings indicate that \textbf{the fragment-level description of the visual explanation and the subsequent summarization of the obtained descriptions is the optimal approach, among the examined ones, for generating plausible textual explanations.}

\subsection{Qualitative results}

\begin{figure*}[t]
\centerline{\includegraphics[width=\textwidth]{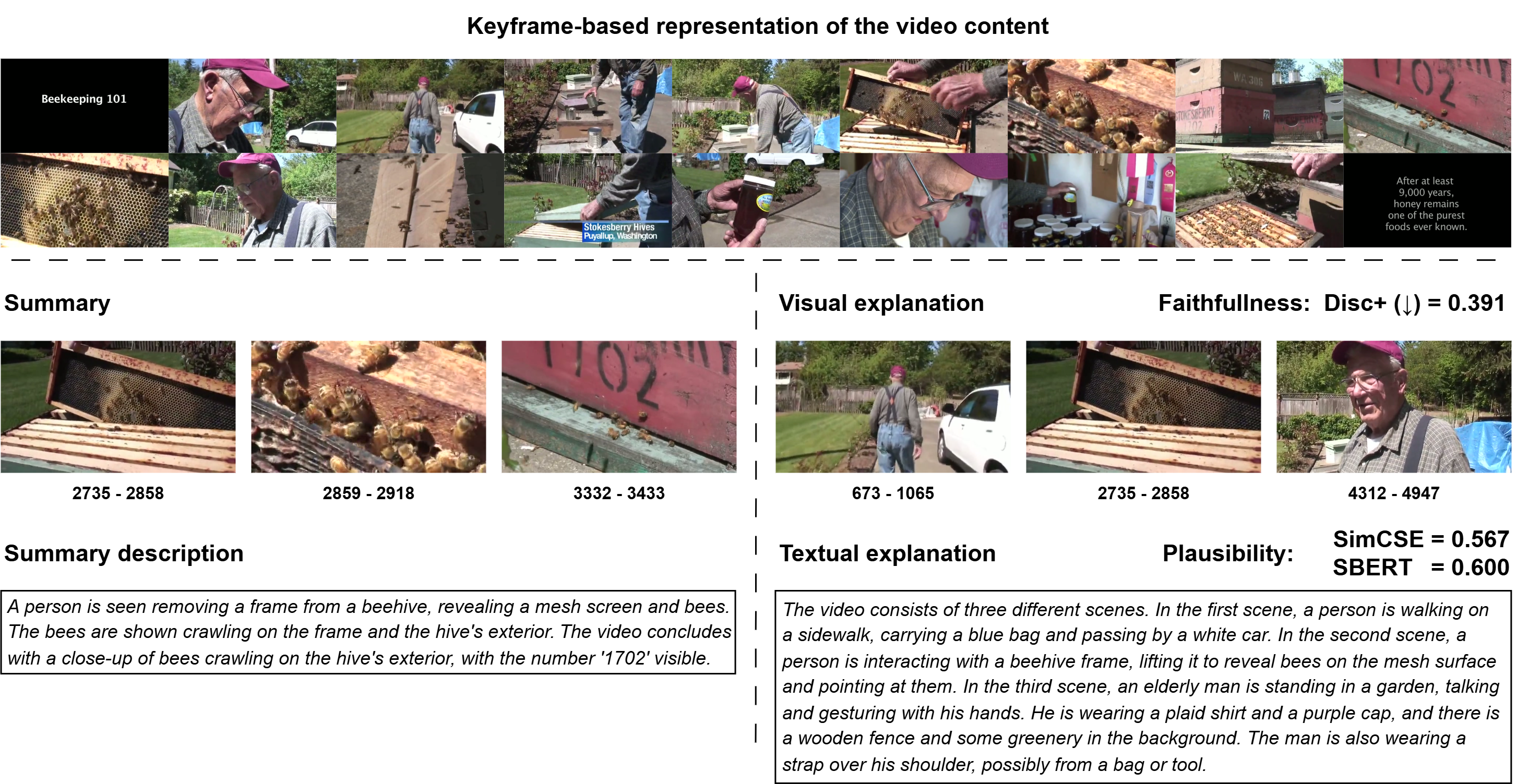}}
\caption{Top: keyframe-based representation of the content of a TVSum video, titled ``Beekeeper''. Middle: keyframe-based representations of the video summary and the produced fragment-level explanation by the attention-based explanation method. Bottom: textual descriptions of the video summary and the fragment-level explanation obtained by the integrated LLaVA-OneVision model, and the computed plausibility scores.}
\label{fig:qual_ex2}
\end{figure*}

Our qualitative analysis was based on the obtained textual explanations for two videos of the TVSum dataset, following the most appropriate approach (Approach 2). The part at the top of Figs. \ref{fig:qual_ex1} and \ref{fig:qual_ex2} provides a keyframe-based representation of the visual content of the full-length video. The part in the middle contains keyframe-based representations of the video summary and the produced fragment-level explanation by the attention-based explanation method. The part at the bottom shows the obtained textual descriptions for the video summary and the fragment-level explanation, as well as the computed plausibility scores.

In the example video of Fig. \ref{fig:qual_ex1}, which is titled ``Reuben Sandwich with Corned Beef \& Sauerkraut'', the generated summary contains parts of the video showing the making of a reuben sandwich. The produced visual explanation shows that the summarization network focuses on the woman in the kitchen presenting the recipe and making the sandwich (first fragment) and the layering of the ingredients and the assembly of the sandwich (second and third fragment). The generated textual explanation provides information about the depicted person and place, and then contains details about the reuben sandwich and parts of the preparation process. The computed plausibility scores indicate that the textual explanation exhibits significant semantic overlap with the content of the video summary - and thus seems reasonable and convincing to human experts - as it focuses on the main objects and activities in the video; i.e., the person who cooks, the cooked meal and the cooking process. 

In the example video of Fig. \ref{fig:qual_ex2}, which is titled ``Beekeeping'', the produced video summary focuses on the beehive and specific actions conducted in the beehive by the beekeeper. The obtained visual explanation indicates that the summarization network paid a lot of attention to the beekeeper, as he appears in two of the three fragments of the explanation. As a consequence, the generated textual explanation presents limited semantic overlap with the content of the video summary, a fact that is indicated also by the significantly lower plausibility scores compared to the ones in our previous example (see Fig. \ref{fig:qual_ex1}). The examples discussed above show that the extended framework for textual explanation of video summarization, allows to produce user-friendly explanations for the outcomes of the video summarization process, that do not require interpretation by human experts. Moreover, the proposed approach for evaluating the plausibility of visual explanations provides reliable cues about their alignment with the humans' reasoning and expectations.

\section{Conclusions}
In this work, we reported our study on generating plausible textual explanations for video summarization. This study was based on the extended version of an existing framework for explainable video summarization, that integrates a Large Multimodal Model for producing natural language descriptions of visual explanations. Building on the obtained descriptions, we proposed to evaluate the plausibility of a visual explanation by quantifying its semantic overlap with the video summary, using a corresponding textual description of the summary and two sentence embedding methods. Our quantitative analysis using a SOTA network architecture and two benchmarking datasets for video summarization, demonstrated that the level of descriptiveness of a visual explanation affects its faithfulness and its alignment with the associated textual explanation in terms of plausibility. Moreover, it allowed to determine the most appropriate approach, among the considered ones, for producing plausible textual explanations. Our qualitative analysis showcased the capacity of the extended framework to produce plausible textual explanations for the outcomes of video summarization.

\balance

\end{document}